\documentclass[twocolumn,10pt]{article}

\usepackage[T1]{fontenc}
\usepackage[utf8]{inputenc}
\usepackage{lmodern}
\usepackage[margin=1in]{geometry}
\usepackage{amsmath,amssymb}
\usepackage{graphicx}
\usepackage{caption}
\usepackage{titlesec}
\usepackage{setspace}
\usepackage{authblk}
\usepackage{fancyhdr}
\usepackage{hyperref}
\usepackage{booktabs}
\usepackage{float}
\usepackage{enumitem}
\usepackage{url}
\usepackage{tabularx}
\usepackage{booktabs}

\titleformat{\section}
  {\bfseries\large}{\thesection.}{0.5em}{}

\titleformat{\subsection}
  {\bfseries}{\thesubsection}{0.5em}{}

\titleformat{\subsubsection}
  {\bfseries\normalsize}{\thesubsubsection}{0.5em}{}

\renewenvironment{abstract}{
\begin{center}
\bfseries ABSTRACT
\end{center}
\itshape
}{\par\vspace{1em}}

\begin{document}

\title{\bfseries How Market Volatility Shapes Algorithmic Collusion: A Comparative Analysis of Learning-Based Pricing Algorithms}

\author[1]{Aheer Sravon}
\author[1]{Md. Ibrahim}
\author[1]{Devdyuti Mazumder}
\author[1]{Ridwan Al Aziz}

\affil[1]{Industrial \& Production Engineering, Bangladesh University of Engineering and Technology, Dhaka, Bangladesh}

\date{} 

\maketitle

\begin{abstract}
Autonomous pricing algorithms are increasingly influencing competition in digital markets; however, their behavior under realistic demand conditions remains largely unexamined. This paper offers a thorough analysis of four pricing algorithms -- Q-Learning, PSO, Double DQN, and DDPG -- across three classic duopoly models (Logit, Hotelling, Linear) and under various demand-shock regimes created by auto-regressive processes. By utilizing profit- and price-based collusion indices, we investigate how the interactions among algorithms, market structure, and stochastic demand collaboratively influence competitive outcomes. Our findings reveal that reinforcement-learning algorithms often sustain supra-competitive prices under stable demand, with DDPG demonstrating the most pronounced collusive tendencies. Demand shocks produce notably varied effects: Logit markets suffer significant performance declines, Hotelling markets remain stable, and Linear markets experience shock-induced profit inflation. Despite marked changes in absolute performance, the relative rankings of the algorithms are consistent across different environments. These results underscore the critical importance of market structure and demand uncertainty in shaping algorithmic competition, while also contributing to the evolving policy discussions surrounding autonomous pricing behavior.
\end{abstract}

\vspace{1em}



\section{Introduction}

The rapid diffusion of autonomous pricing algorithms has reshaped competitive 
dynamics in digital marketplaces, raising important economic and policy 
questions about their potential for collusive behavior. A substantial body of 
research demonstrates that reinforcement-learning (RL) agents can autonomously 
coordinate on supracompetitive outcomes even in the absence of explicit 
communication. Foundational contributions---including the work in \hyperlink{ref1}{\textbf{[1]}}---show 
that algorithmic agents may systematically learn tacitly collusive strategies 
across multiple market structures, with Q-learning in particular generating 
prices above competitive levels in Logit, Hotelling, and linear demand 
environments. These concerns are reinforced by seminal work such as \hyperlink{ref2}{\textbf{[2]}}, 
which demonstrates that simple Q-learning agents reliably sustain collusion 
through structured punishment and reward cycles in repeated pricing games, as 
well as by \hyperlink{ref3}{\textbf{[3]}}, who document how algorithmic systems may generate sudden 
price spikes in response to high-impact, low-probability events (HILP), 
unintentionally coordinating on elevated prices. The study of \hyperlink{ref4}{\textbf{[4]}} 
establishes a robust empirical and computational foundation demonstrating that 
pricing algorithms may autonomously learn to collude.

A complementary line of research focuses specifically on Q-learning's 
capacity to learn collusive equilibria, as documented in papers \hyperlink{ref2}{\textbf{[2]}}, 
\hyperlink{ref5}{\textbf{[5]}}, and \hyperlink{ref6}{\textbf{[6]}}. These findings are consistent with the theoretical 
properties of Q-learning established by \hyperlink{ref7}{\textbf{[7]}}, who show that the algorithm 
incrementally learns long-run discounted value-maximizing strategies in 
sequential decision problems. More recent studies further reveal that deep 
reinforcement-learning (deep RL) algorithms---including DDQN and SAC---may also 
display collusive tendencies. For instance, \hyperlink{ref8}{\textbf{[8]}} documents that modern RL 
systems can coordinate on higher-than-competitive prices under a variety of 
market configurations.

Despite these advances, the existing literature remains narrow in scope. The 
paper \hyperlink{ref3}{\textbf{[3]}} considers competition only between Q-learning and Particle 
Swarm Optimization (PSO) across three canonical market structures (Logit, 
Hotelling, and linear). The Logit model has become a standard in the algorithmic 
pricing literature \hyperlink{ref2}{\textbf{[2]}}, \hyperlink{ref5}{\textbf{[5]}}, \hyperlink{ref6}{\textbf{[6]}}. The Hotelling and the linear demand model 
are well-known in operations research and economic modeling \hyperlink{ref9}{\textbf{[9]}}, \hyperlink{ref10}{\textbf{[10]}}.

Nevertheless, contemporary pricing systems increasingly employ deep RL 
architectures, notably Double Deep Q-Networks(DDQN) and Deep Deterministic Policy 
Gradient (DDPG), which allow firms to process high-dimensional state spaces and 
adjust prices over continuous action domains \hyperlink{ref11}{\textbf{[11]}}. DQN, first introduced 
by \hyperlink{ref12}{\textbf{[12]}} and later stabilized through the Double-DQN variant proposed by 
\hyperlink{ref13}{\textbf{[13]}}, leverages experience replay and target networks to approximate 
value functions using deep neural networks. DDPG, introduced by \hyperlink{ref14}{\textbf{[14]}}, 
extends these methods to continuous action spaces through an actor--critic 
architecture, making it highly suitable for pricing environments where price 
adjustments are granular and continuous. Subsequent applications in pricing, 
control, and energy markets further demonstrate DDPG's stability and 
scalability \hyperlink{ref15}{\textbf{[15]}}, \hyperlink{ref16}{\textbf{[16]}}. However, despite their growing real-world 
relevance, the competitive behavior of these deep RL algorithms remains 
insufficiently explored.

A second limitation of the existing body of work is the near-universal use of 
deterministic demand. Fundamental markets, however, are inherently stochastic, 
subject to temporal fluctuations, shocks, and random disturbances \hyperlink{ref17}{\textbf{[17]}}, \hyperlink{ref18}{\textbf{[18]}}. 
The literature provides numerous models for representing these shocks, including 
Auto Regressive Shock of Order 1 (AR(1)) processes \hyperlink{ref19}{\textbf{[19]}}, Hotelling-based randomness, Logit-type shocks \hyperlink{ref20}{\textbf{[20]}}, 
and linear demand perturbations, including multiple variants. However, the 
interaction between deep RL algorithms and stochastic demand remains largely 
unexamined, despite the fact that randomness may either disrupt or amplify 
collusive equilibria.

This paper addresses these gaps by extending the framework. First, we reproduce 
the original competitive environment, validating the behavior of Q-learning and 
PSO in Logit, Hotelling, and linear demand duopoly markets. Second, we introduce 
two advanced deep RL algorithms---DQN and DDPG---grounded in the foundational work 
of \hyperlink{ref12}{\textbf{[12]}}, \hyperlink{ref13}{\textbf{[13]}}, \hyperlink{ref14}{\textbf{[14]}}, \hyperlink{ref16}{\textbf{[16]}}. Third, we incorporate stochastic demand shocks using an AR(1) process; 
these shocks are unobservable to firms and enter additively as latent 
demand disturbances that evolve over time \hyperlink{ref19}{\textbf{[19]}}, \hyperlink{ref21}{\textbf{[21]}}. Fourth, we 
systematically evaluate how algorithmic behavior varies across market structures 
and shock environments, with particular attention to collusion indicators such 
as price stability, deviation--punishment cycles, and convergence toward 
supracompetitive equilibria \hyperlink{ref22}{\textbf{[22]}}.

By expanding both the algorithmic scope and the market complexity of the 
benchmark model, this study offers the most comprehensive analysis to date of 
how modern RL algorithms behave under competitive pressure, particularly in the 
presence of stochastic demand. Our findings provide insights relevant for 
economists, computer scientists, policymakers, and practitioners designing or 
regulating algorithmic pricing systems.

\section{Methodology}

\subsection{Market Environment}

We study a symmetric duopoly with differentiated products in three standard demand environments: Logit, Hotelling, and Linear. In each instance, the demand system and parameterization adhere to the baseline calibration established by \hyperlink{ref1}{\textbf{[1]}}, drawing on functional forms rooted in the industrial organization literature. Time is discrete, $t = 1,\dots,T$, with horizon $T = 10{,}000$.

In all cases, each firm chooses a price from a discrete grid of 
$N = 15$ points constructed around the static Nash and monopoly prices 
$P_{N}$ and $P_{M}$ of the corresponding deterministic model. 
The grid is
\[
P = \{p_{1},\ldots,p_{15}\},
\]
where 

\begin{equation}
\begin{aligned}
p_{1}  &= P_{N} - 0.15\,(P_{M} - P_{N}), \\
p_{15} &= P_{M} + 0.15 + \,(P_{M} - P_{N}).
\end{aligned}
\end{equation}

and the remaining prices $p_{2},\ldots,p_{14}$ are evenly (linearly) 
spaced between $p_{1}$ and $p_{15}$.

Per-period profit of firm $i$ is
\begin{equation}
\pi_i(p_i,p_j) = (p_i - c)\, q_i(p_i,p_j).
\end{equation}

\subsubsection{Logit demand}

Consumers choose between the two inside goods and an outside option. Deterministic utilities are
\begin{equation}
u_i = a - \mu p_i, 
\qquad
u_0 = a_0.
\end{equation}

Market shares follow the multinomial Logit form:
\begin{equation}
\begin{aligned}
s_i(p_i,p_j) = 
\frac{e^{u_i}}{e^{u_1} + e^{u_2} + e^{u_0}}, \\
s_0 = 
\frac{e^{u_0}}{e^{u_1} + e^{u_2} + e^{u_0}}.
\end{aligned}
\end{equation}

and demand for firm $i$ is $q_i = s_i$ (unit mass of consumers).

We use the calibration implemented in the environment:
\[
a = 2.0,\quad a_0 = 0.0,\quad \mu = 0.25,\quad c = 1.0.
\]
For this parametrization, the static benchmarks are $P_N = 1.473$ and $P_M = 1.925$, which are hard-coded and used to build the price grid. This specification corresponds to the standard discrete-choice Logit model with an outside good as presented in \hyperlink{ref23}{\textbf{[23]}} and is widely used in the algorithmic pricing literature \hyperlink{ref2}{\textbf{[2]}} and \hyperlink{ref24}{\textbf{[24]}}.

\subsubsection{Hotelling demand}

Firms are located at $0$ and $1$ on the unit interval; consumers are uniformly distributed on $[0,1]$. Utility of a consumer at $x$ from firm $i$ is
\begin{equation}
U_i(x) = v - p_i - \theta\, |x - x_i|.
\end{equation}

The indifferent consumer $\hat{x}$ solves $U_1(\hat{x}) = U_2(\hat{x})$, and demands are
\begin{equation}
q_1 = \hat{x}, \qquad q_2 = 1 - \hat{x}.
\end{equation}

We use a symmetric full-coverage calibration:
\[
v = 1.75,\quad \theta = 1.0,\quad c = 0.0,
\]

In the deterministic model, the environment fixes $P_N = 1.00$ and $P_M = 1.25$. This is the standard linear Hotelling duopoly as in \hyperlink{ref9}{\textbf{[9]}} and \hyperlink{ref10}{\textbf{[10]}}. 

\subsubsection{Linear demand}

The linear differentiated-products environment is
\begin{equation}
q_i = a - p_i - d.
\end{equation}
Solving the two-equation system yields

\begin{equation}
\begin{aligned}
q_1 = \frac{(a - p_1) - d(a - p_2)}{1 - d^2}, \\
q_2 = \frac{(a - p_2) - d(a - p_1)}{1 - d^2}.
\end{aligned}
\end{equation}

We use the calibration
\[
a = \bar{a} = 1.0,\quad d = 0.25,\quad c = 0.0.
\]
With these parameters, the environment's initialization computes and stores the no-shock benchmarks $P_N \approx 0.4286$ and $P_M = 0.5$, together with the corresponding Nash and monopoly profits, for later use in collusion measures. This linear specification corresponds to the classical Bertrand model with differentiated products used in \hyperlink{ref9}{\textbf{[9]}} and \hyperlink{ref10}{\textbf{[10]}}.

\subsection{Demand Shocks}

To introduce stochastic demand, we superimpose additive shocks on the underlying demand system via an autoregressive AR(1) process \hyperlink{ref10}{\textbf{[10]}}, \hyperlink{ref21}{\textbf{[21]}}.

\subsubsection{Auto Regressive Shock of Order 1(AR(1)) process}

For each firm $i$, the shock $z_{i,t}$ evolves as
\begin{equation}
z_{i,t} = \rho\, z_{i,t-1} + \eta_{i,t}, 
\qquad 
\eta_{i,t} \sim \mathcal{N}(0,\sigma_\eta^2).
\end{equation}
In the simulations with shocks, firms face independent AR(1) processes (separate draws for each firm). The environment also supports a correlated mode, but we do not use it in the main experiments reported here. 

\subsubsection{Mapping shocks into demand}

Shocks are integrated into each demand system as follows.

\paragraph{Linear.} Shocks perturb the intercept:
\begin{equation}
a_t = \bar{a} + z_t, 
\qquad 
q_i = a_t - p_i - d.
\end{equation}

\paragraph{Logit.} Shocks perturb deterministic utility:
\begin{equation}
u_i = a - \mu p_i + z_{i,t},
\end{equation}
so they act as time-varying quality/taste shifts.

\paragraph{Hotelling.} Shocks shift effective valuations $v$, thereby moving the indifferent consumer and hence changing $q_i$.

Because the Logit model is nonlinear, shocks change expected Nash and monopoly benchmarks; for Hotelling and Linear, expected benchmarks coincide with the no-shock case.

\subsubsection{Shock regimes}

We consider four regimes:

\begin{table}[h!]
\centering
\caption{Shock regimes and AR(1) parameter values.}
\begin{tabularx}{0.48\textwidth}{lccX}
\toprule
\textbf{Regime} & $\boldsymbol{\rho}$ & $\boldsymbol{\sigma_\eta}$ & \textbf{Description} \\
\midrule

No-shock baseline & 0 & 0 
& Shocks disabled: $z_{i,t} \equiv 0$. \\

Scheme A & 0.3 & 0.5
& Low persistence, high volatility. \\

Scheme B & 0.95 & 0.05
& Near-unit persistence, low volatility. \\

Scheme C & 0.9 & 0.3
& High persistence, high volatility. \\

\bottomrule
\end{tabularx}

\end{table}

The same scheme parameters are used both in the environment and in the theoretical benchmark calculations.

\subsection{Algorithms and State/Action Spaces}

We compare four algorithms:
Q-Learning,
Particle Swarm Optimization (PSO),
Double DQN,
Deep Deterministic Policy Gradient (DDPG).

\subsection{Q-Learning}

Each Q-learning firm maintains a Q-table $Q_i(s,a)$ of dimension $225 \times 15$ (opponent's index $\times$ own action), initialized at zero \hyperlink{ref7}{\textbf{[7]}}. After choosing action $a$ in state $s$, observing profit $\pi_i$ and next state $s'$, the update is
\begin{equation}
Q_i(s,a) \leftarrow Q_i(s,a) 
+ \alpha\Bigl[ \pi_i + \delta \max_{a'} Q_i(s',a') - Q_i(s,a)\Bigr].
\end{equation}

The hyperparameters for the Q-learning agent are as follows. The learning rate is $\alpha = 0.15$ and the discount factor is $\delta = 0.95$. Action selection follows an $\varepsilon$-greedy strategy with exponentially decaying exploration, where $\varepsilon_t = \exp(-\beta t)$ and $\beta = 1.5 \times 10^{-4}$. With probability $\varepsilon_t$ the agent selects a random action, while with probability $1 - \varepsilon_t$ it chooses the greedy action $\arg\max_a Q_i(s,a)$.

\subsection{Particle Swarm Optimization (PSO)}

PSO provides a search-based benchmark without temporal learning. Each firm controls a swarm of 10 particles $x^\ell_t \in [0,2]$ with associated velocities $v^\ell_t$. Given the opponent's current price $p^j_t$, each particle's price $x^\ell_t$ is evaluated via the environment to obtain profit $\pi_i(x^\ell_t,p^j_t)$. Personal best positions $p_{\text{best},\ell}$ and the global best $g_{\text{best}}$ are updated accordingly.

Velocities are updated as
\begin{equation}
v^\ell_{t+1} = w_t v^\ell_t 
+ c_1 r_1 \bigl(p_{\text{best},\ell} - x^\ell_t\bigr)
+ c_2 r_2 \bigl(g_{\text{best}} - x^\ell_t\bigr),
\end{equation}
with:
\[
w_t = \max\bigl\{0.4,\, 0.9 - 0.5\, t / 10{,}000\bigr\} \quad \text{(inertia weight)},
\]
\[
c_1 = c_2 = 1.75 \quad \text{(cognitive and social coefficients)},
\]
\[
r_1,r_2 \sim U[0,1],
\]
and velocities clipped to $[-0.3,0.3]$.

Positions are then updated and clipped:
\begin{equation}
x^\ell_{t+1} = \mathrm{clip}\bigl(x^\ell_t + v^\ell_{t+1},\, 0,\, 2\bigr).
\end{equation}
The firm's actual price in period $t$ is the current global best $g_{\text{best},t}$. A restart mechanism reinitializes the swarm if the global best does not improve for 300 consecutive steps, to avoid stagnation in poor local optima.

\subsection{Double DQN}

The Double DQN agent approximates the action-value function $Q_\theta(s,a)$ using a neural network \hyperlink{ref11}{\textbf{[11]}}, \hyperlink{ref12}{\textbf{[12]}}, \hyperlink{ref13}{\textbf{[13]}}. The state is $s_t = (p^i_t, p^j_t)$, the pair of current prices, and the action $a_t$ is one of the 15 discrete price indices.

Given a transition $(s_t,a_t,r_t,s_{t+1})$, Double DQN mitigates the upward bias of standard DQN by decoupling action selection and evaluation. The online network $Q_\theta$ selects the greedy action at the next state, while the target network $Q_{\theta^-}$ evaluates it:
\begin{equation}
y_t = r_t + \gamma\, Q_{\theta^-}\!\left(s_{t+1},\, \arg\max_{a'} Q_\theta(s_{t+1},a')\right),
\end{equation}
where $r_t$ is the observed profit and $\gamma$ the discount factor. The online network minimizes
\begin{equation}
L(\theta) = \bigl(y_t - Q_\theta(s_t,a_t)\bigr)^2,
\end{equation}
and the target network is updated every 500 gradient steps. Here, $Q_\theta$ is the online value estimate, $Q_{\theta^-}$ the stable target estimate, and $\arg\max_{a'} Q_\theta(s_{t+1},a')$ selects the next greedy action.

The neural network has architecture $2 \to 128 \to 128 \to 64 \to 15$, mapping the price pair to Q-values for each action. Hidden layers use ReLU activations, and training relies on experience replay with periodic target-network updates.

Training uses a discount factor $\gamma = 0.99$, an Adam optimizer with learning rate $10^{-4}$, a replay buffer of $50{,}000$ transitions, minibatch size 128, gradient clipping (norm~1.0), and the Huber loss. Exploration follows an $\varepsilon$-greedy schedule with $\varepsilon_0 = 1.0$, decaying as $\varepsilon \leftarrow \max\{\varepsilon_{\min},\, 0.995\,\varepsilon\}$ and $\varepsilon_{\min} = 0.01$.

\subsection{Deep Deterministic Policy Gradient (DDPG)}

DDPG learns a continuous pricing policy with separate actor and critic networks. The state is $s_t = (p^i_t,p^j_t)$ (current prices of the two firms), and the actor outputs a normalized action $a_t \in [-1,1]$, which is mapped to a price via $p^i_t = 1 + a_t$ given $[p_{\min},p_{\max}] = [0,2]$.

The actor network is a fully connected architecture $2 \to 400 \to 300 \to 1$ with Rectified Linear Unit (ReLU) activations in the hidden layers and a $\tanh$ output to enforce the $[-1,1]$ bound. The critic takes the state as input, applies batch normalization, passes it through a 400-unit ReLU layer, concatenates the action, and applies a 300-unit ReLU layer before producing the scalar estimate $Q(s,a)$.

Learning follows a deterministic actor - critic scheme based on temporal-difference (TD) updates. The critic minimizes a TD loss, i.e., the squared deviation between a target value computed with the target critic and the current estimate $Q(s_t,a_t)$, while the actor is updated to maximize $Q(s,a)$ with respect to its parameters. Let $\theta$ and $\theta^-$ denote the parameter vectors of an online network (actor or critic) and its corresponding target network, and let $\tau \in (0,1)$ be the soft-update coefficient. Target parameters are updated according to
\[
\theta^- \leftarrow \tau \theta + (1 - \tau)\theta^-,
\]
so that $\theta^-$ tracks $\theta$ as an exponential moving average.

The main hyperparameters are: discount factor $\gamma = 0.99$; Adam optimizer with learning rate $10^{-4}$ for the actor and $10^{-3}$ for the critic (with $\ell_2$ weight decay $10^{-2}$); replay buffer capacity of $1{,}000{,}000$ transitions; and minibatch size of 64.

Exploration noise is generated by an Ornstein--Uhlenbeck process. Let $\xi_t$ denote the exploration noise, with long-run mean $\mu$, mean-reversion rate $\theta$ (here a noise parameter), volatility $\sigma$, and $\varepsilon_t$ an i.i.d.\ standard normal shock. The process evolves as
\begin{equation}
\xi_{t+1} = \xi_t + \theta(\mu - \xi_t) + \sigma \varepsilon_t,
\end{equation}
and the resulting perturbation is scaled by an external exploration parameter that decays from $1.0$ to $0.01$ at rate $0.995$ \hyperlink{ref8}{\textbf{[8]}}, \hyperlink{ref14}{\textbf{[14]}}.

\subsection{Benchmarks and Collusion Measures}

For each market and shock configuration, we compute the Nash and monopoly benchmarks $(p_N, p_M, \pi_N, \pi_M)$, analytically for the Hotelling and Linear models and through numerical optimization with Monte Carlo integration for the Logit model with shocks. These serve as reference points for normalizing simulated prices and profits. To evaluate long-run behavior, we use each agent's average price $\bar{p}_i$ and profit $\bar{\pi}_i$ over the final 1{,}000 periods of each run.

We employ two standard collusion indicators that capture different dimensions of strategic behavior. The profit-based collusion index, introduced in \hyperlink{ref2}{\textbf{[2]}}, compares realized profits with the Nash and monopoly benchmarks. Let $\bar{\pi}_i$ be the average profit of firm $i$; then
\begin{equation}
    (\Delta_i) = \frac{\bar{\pi}_i - \pi_N}{\pi_M - \pi_N},
\end{equation}
so that $\Delta_i = 0$ corresponds to Nash profits and $\Delta_i = 1$ corresponds to monopoly profits. Values may exceed one when profits surpass the benchmark monopoly level.

The price-based collusion index follows \hyperlink{ref22}{\textbf{[22]}} and measures how observed prices compare to the Nash - monopoly interval. Let $\bar{p}_i$ denote firm $i$'s average price; then
\begin{equation}
\mathrm{RPDI}_i = \frac{\bar{p}_i - p_N}{p_M - p_N},
\end{equation}
with $0$ indicating Nash pricing and $1$ indicating monopoly pricing. Values above one reflect prices that exceed the monopoly benchmark. Because prices and profits need not align in asymmetric or unstable environments, the two indicators may diverge: high prices do not guarantee high profits, and moderate pricing may still yield high profits depending on competitor behavior. Reporting both indices provides a clearer picture of algorithmic conduct across treatments.

\subsection{Experimental Design}

The experimental design evaluates the four algorithms -- Q-Learning, PSO, Double DQN, and DDPG -- by allowing each algorithm to interact with each other, including itself. Each pair of algorithms is tested across the three market structures (Logit, Hotelling, and Linear) and under the four demand-shock regimes (no shocks, Scheme A, Scheme B, and Scheme C), yielding twelve environment configurations for every algorithmic pairing.

For each configuration, simulations run for $T = 10{,}000$ periods, and multiple independent runs are performed under different random seeds to account for stochastic variation in demand shocks and learning dynamics. In every run, prices and profits are recorded each period, and long-run behavior is measured using the final 1{,}000 periods. For each firm, the collusion measures $\Delta_i$ and $\mathrm{RPDI}_i$ are computed from these terminal observations and then averaged across simulation replications.

This design provides a systematic comparison of algorithmic behavior across alternative learning rules, market structures, and stochastic environments, ensuring that conclusions reflect both cross-algorithm robustness and the influence of demand uncertainty.

\section{Results}

\subsection{Baseline Results: Homogeneous Competition Without Demand Shocks}

\begin{figure*}
\centering
\includegraphics[width=0.9\textwidth]{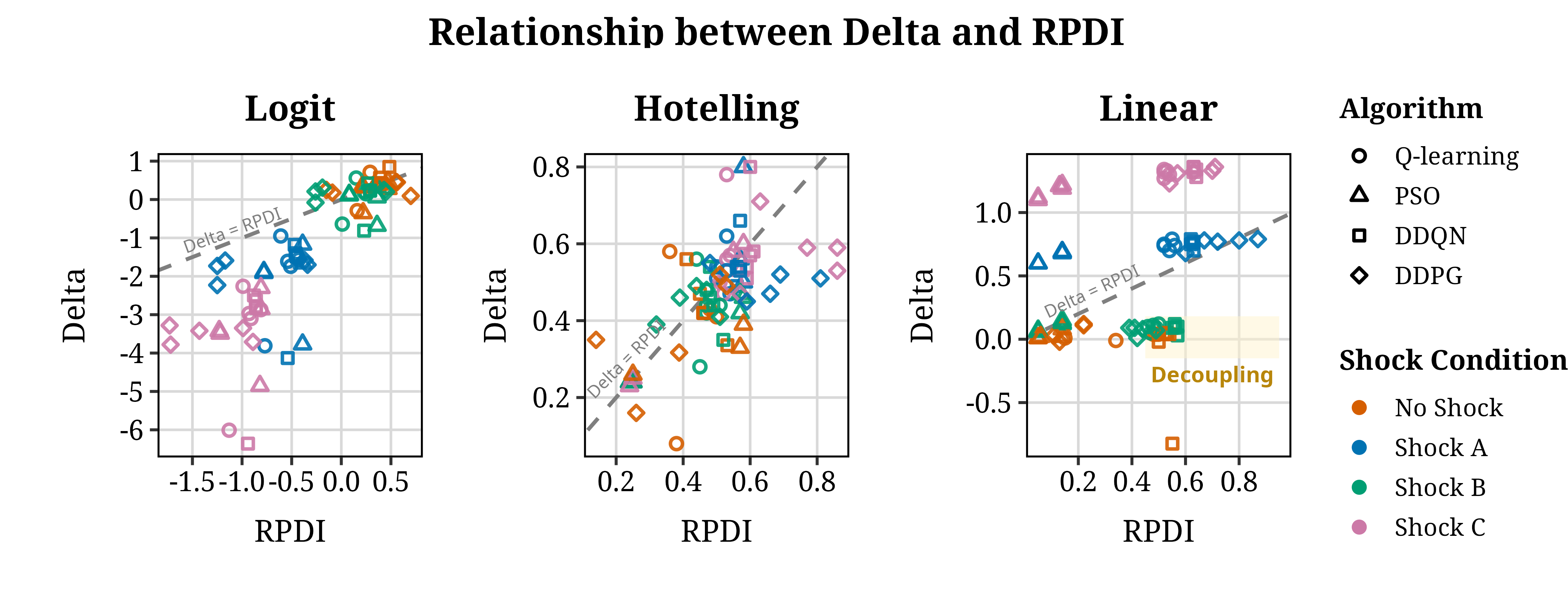}
\caption{Delta vs. RPDI Scatter Plots by Algorithm}
\label{fig:delta_rpdi}
\end{figure*}

Table~\ref{tab:baseline_collusion} reports the mean values of the profit-based indicator ($\Delta$) 
and the price-based indicator (RPDI) for each algorithm competing against itself across the three 
market structures. Figure~\ref{fig:delta_rpdi} complements these baseline values by illustrating 
the full distribution of $(\Delta, \text{RPDI})$ pairs across all simulation runs and shock regimes.

Table~\ref{tab:baseline_collusion} shows that algorithmic collusion is not an inherent attribute of specific learning architectures but emerges from the interaction between algorithm design and market structure. Three central patterns arise.

\begin{table*}
\centering
\caption{Collusion Indicators Under Homogeneous Competition (No Shocks)}
\label{tab:baseline_collusion}
\begin{tabular}{lcccccc}
\toprule
 & \multicolumn{2}{c}{\textbf{Logit}} & \multicolumn{2}{c}{\textbf{Hotelling}} & \multicolumn{2}{c}{\textbf{Linear}} \\
\cmidrule(lr){2-3} \cmidrule(lr){4-5} \cmidrule(lr){6-7}
\textbf{Algorithm} & $\mathbf{\Delta}$ & \textbf{RPDI} & $\mathbf{\Delta}$ & \textbf{RPDI} & $\mathbf{\Delta}$ & \textbf{RPDI} \\
\midrule
\textbf{Q-Learning} & 0.36 & 0.36 & 0.44 & 0.48 & 0.04 & 0.49 \\
\textbf{DDQN} & 0.42 & 0.41 & 0.42 & 0.46 & 0.05 & 0.54 \\
\textbf{PSO} & 0.33 & 0.22 & 0.24 & 0.25 & 0.01 & 0.05 \\
\textbf{DDPG} & 0.46 & 0.56 & 0.51 & 0.52 & 0.12 & 0.22 \\
\bottomrule
\end{tabular}
\end{table*}

\textbf{First}, collusive propensities are highly market dependent. In both the Logit and Hotelling environments, the deep reinforcement learning algorithms -- DDQN and DDPG -- systematically generate the strongest signs of tacit coordination, with jointly elevated profit-based ($\Delta$) and price-based (RPDI) indicators. DDPG, in particular, exhibits the highest RPDI in the Logit model (0.56), consistent with prior findings on policy-based methods in differentiated markets \hyperlink{ref29}{\textbf{[29]}}, \hyperlink{ref30}{\textbf{[30]}}.

\textbf{Second}, the Linear model produces a qualitatively distinct pattern. Here, Q-Learning and DDQN sustain higher-than-competitive prices (RPDI of 0.49 and 0.54) while earning nearly competitive profits ($\Delta$ of 0.04 and 0.05). This ``decoupling'' pattern -- prices rising without corresponding profit gains -- suggests that consumer surplus declines while firms capture little benefit. The mechanism stems from the high elasticity of linear demand, where aggressive price increases trigger substantial quantity losses, reinforcing that algorithmic outcomes are deeply conditioned by the structure of demand \hyperlink{ref1}{\textbf{[1]}}, \hyperlink{ref2}{\textbf{[2]}}.

\textbf{Third}, PSO consistently converges to near-competitive outcomes across all market structures, with both indicators remaining close to zero. Its behavior reflects the absence of temporal reasoning and memory, features that enable RL-based agents to sustain tacit coordination \hyperlink{ref31}{\textbf{[31]}}. In contrast, DDPG exhibits tightly aligned $\Delta$ and RPDI values in the Hotelling market (0.51, 0.52), indicating stable supra-competitive pricing in spatially differentiated settings -- a pattern consistent with the strategic flexibility of actor--critic architectures \hyperlink{ref30}{\textbf{[30]}}, \hyperlink{ref31}{\textbf{[32]}}.

\subsubsection{Algorithm-Specific Analysis}

\textbf{Q-Learning} exhibits a clear tendency toward supra-competitive pricing, though its profitability varies substantially across market structures. In the Hotelling model, it records the highest profit-based indicator ($\Delta=0.44$) among all algorithms, reflecting a capacity for tacit coordination in differentiated markets. This pattern is consistent with established literature documenting Q-Learning agents' development of coordinated pricing strategies in repeated interactions without explicit communication \hyperlink{ref2}{\textbf{[2]}}, \hyperlink{ref5}{\textbf{[5]}}. However, the Linear model reveals a distinct outcome, where Q-Learning maintains elevated prices (RPDI=0.49) while capturing near-competitive profits ($\Delta=0.04$). This decoupling indicates that prices rise without proportional profit increases, reducing consumer surplus while providing minimal gains to firms---a pattern likely attributable to the model's elastic demand structure that penalizes price increases through substantial market share losses \hyperlink{ref1}{\textbf{[1]}}.

\textbf{PSO} demonstrates outcomes closest to competitive equilibrium, consistently converging to near-Nash equilibrium across all market structures. It records the lowest values for both indicators in the Linear model ($\Delta=0.01$, RPDI=0.05) and maintains near-competitive pricing and profits in Logit and Hotelling models. This pattern aligns with literature findings that PSO converges to equilibrium prices and often produces outcomes associated with higher social welfare compared to reinforcement learning algorithms \hyperlink{ref1}{\textbf{[1]}}, \hyperlink{ref33}{\textbf{[33]}}. These results suggest that PSO exhibits lower propensity for supra-competitive pricing, representing a distinct behavioral profile in algorithmic pricing contexts.

\textbf{DDQN} demonstrates a distinct competitive profile across market structures. While literature suggests it can converge to more competitive prices than Q-learning by mitigating overestimation bias \hyperlink{ref11}{\textbf{[11]}}, \hyperlink{ref34}{\textbf{[34]}}, our results indicate it maintains substantial supra-competitive pricing. In the Hotelling and Linear models, DDQN's RPDI (0.46 and 0.54) is among the highest observed, reflecting prices considerably above Nash equilibrium. This is particularly evident in the Linear model, where it records the highest RPDI of any algorithm while capturing minimal profits ($\Delta=0.05$), producing a decoupling between price elevation and profit capture. This pattern suggests that while DDQN's technical enhancements may improve learning stability and profit outcomes \hyperlink{ref35}{\textbf{[35]}}, they do not inherently prevent supra-competitive pricing behavior, particularly in markets where demand elasticity provides limited constraint on price increases.

\textbf{DDPG} demonstrates consistent supra-competitive pricing patterns across markets, recording the highest profit-based indicators ($\Delta$) in both Logit (0.46) and Hotelling (0.51) models. Its continuous action space architecture enables precise price adjustments, which in this context facilitates convergence to and maintenance of supra-competitive equilibria \hyperlink{ref29}{\textbf{[29]}}, \hyperlink{ref36}{\textbf{[36]}}. This is further reflected in its elevated RPDI values, particularly in the Logit model (0.56), indicating both substantial profit extraction and sustained price elevation. The actor-critic architecture enables DDPG to calibrate prices that maximize joint profits while avoiding price wars \hyperlink{ref31}{\textbf{[31]}}. These characteristics indicate that DDPG exhibits pronounced departures from competitive equilibrium, representing a distinct behavioral profile in algorithmic pricing contexts with implications for market outcomes and consumer surplus.

\begin{table}[H]
\centering
\caption{Price-Profit Efficiency Ratio ($\Delta$/RPDI)}
\label{tab:Price-Profit Efficiency Ratio}
\begin{tabular}{lccc}
\toprule
\textbf{Algorithm} & \textbf{Logit} & \textbf{Hotelling} & \textbf{Linear} \\
\midrule
\textbf{Q-Learning} & 1.00 & 0.92 & 0.08 \\
\textbf{DDQN} & 1.02 & 0.91 & 0.09 \\
\textbf{PSO} & 1.50 & 0.96 & 0.20 \\
\textbf{DDPG} & 0.82 & 0.98 & 0.55 \\
\bottomrule
\end{tabular}
\end{table}

Table~\ref{tab:Price-Profit Efficiency Ratio} quantifies the price-profit efficiency across algorithms and market structures. PSO exhibits the highest ratio in the Logit model (1.50), while DDPG records the highest in the Linear model (0.55). Q-Learning and DDQN display notably low ratios in the Linear model, where elevated prices (RPDI $>$ 0.48) correspond to near-competitive profits ($\Delta <$ 0.05). This decoupling indicates that price increases do not translate into proportional profit gains, reducing consumer surplus while providing minimal benefits to firms. The Hotelling model shows ratios approaching unity for most algorithms, suggesting that its spatial differentiation structure facilitates a more direct correspondence between pricing and profitability outcomes.

\subsection{Heterogeneous Competition: Cross-Algorithm Matchups}

Real-world digital markets rarely consist of homogeneous pricing technologies; rather, firms deploy heterogeneous algorithms that differ in learning architecture, state representation, and temporal reasoning. Table~\ref{tab:Heterogeneous Competition Results} summarizes the resulting outcomes across all pairwise algorithmic matchups. Three core patterns emerge.

First, heterogeneous competition produces pronounced asymmetries in profit extraction. In the Logit model, \textit{Q-Learning vs.\ PSO} illustrates this clearly: Q-Learning earns more than twice the profit of PSO ($\Delta = 0.71$ vs.\ 0.28) despite PSO sustaining substantially higher prices (RPDI = 0.47 vs.\ 0.29). This divergence reflects differences in learning dynamics. Q-Learning's temporal-difference updates allow it to interpret PSO's consistently elevated prices as a stable signal, enabling systematic undercutting while maintaining margins. By contrast, PSO's swarm-based heuristic lacks strategic memory and cannot develop adaptive counter-responses, consistent with earlier findings on algorithmic adaptivity \hyperlink{ref1}{\textbf{[1]}}, \hyperlink{ref30}{\textbf{[30]}}, \hyperlink{ref31}{\textbf{[31]}}.

\begin{table*}
\centering
\caption{Heterogeneous Competition Results}
\label{tab:Heterogeneous Competition Results}
\begin{tabular}{lcccccc}
\toprule
 & \multicolumn{2}{c}{\textbf{Logit Model}} & \multicolumn{2}{c}{\textbf{Hotelling Model}} & \multicolumn{2}{c}{\textbf{Linear Model}} \\
\cmidrule(lr){2-3} \cmidrule(lr){4-5} \cmidrule(lr){6-7}
\textbf{Matchup} & $\mathbf{\Delta}$ & \textbf{RPDI} & $\mathbf{\Delta}$ & \textbf{RPDI} & $\mathbf{\Delta}$ & \textbf{RPDI} \\
\midrule
\textbf{Q-Learning vs PSO} & 0.71 & 0.29 & 0.58 & 0.36 & 0.01 & 0.15 \\
 & 0.28 & 0.47 & 0.33 & 0.57 & 0.02 & 0.14 \\
\midrule
\textbf{Q-Learning vs DDQN} & 0.40 & 0.41 & 0.41 & 0.50 & 0.04 & 0.50 \\
 & 0.43 & 0.40 & 0.47 & 0.45 & 0.04 & 0.52 \\
\midrule
\textbf{Q-Learning vs DDPG} & -0.29 & 0.16 & 0.08 & 0.38 & -0.01 & 0.34 \\
 & 0.18 & -0.09 & 0.35 & 0.14 & 0.03 & 0.11 \\
\midrule
\textbf{PSO vs DDQN} & 0.47 & 0.47 & 0.39 & 0.58 & 0.08 & 0.14 \\
 & 0.57 & 0.39 & 0.56 & 0.41 & -0.02 & 0.50 \\
\midrule
\textbf{PSO vs DDPG} & -0.35 & 0.22 & 0.26 & 0.25 & 0.02 & 0.06 \\
 & 0.25 & -0.15 & 0.16 & 0.26 & -0.02 & 0.13 \\
\midrule
\textbf{DDQN vs DDPG} & 0.85 & 0.48 & 0.34 & 0.53 & -0.82 & 0.55 \\
 & 0.09 & 0.70 & 0.32 & 0.39 & -13.45 & -0.10 \\
\bottomrule
\end{tabular}

\end{table*}
Second, DDPG demonstrates a distinctive profit - price relationship in differentiated environments. Against Q-Learning in the Hotelling model, DDPG generates disproportionate profits relative to its price increase ($\Delta = 0.35$, RPDI = 0.14), achieving nearly 2.5 times more profit per unit of price elevation. This stems from continuous action refinement within its actor - critic architecture, allowing highly granular price adjustments that exploit spatial differentiation while limiting provocations that trigger aggressive responses. The mechanism aligns with observations on policy-gradient methods in differentiated markets \hyperlink{ref29}{\textbf{[29]}}, \hyperlink{ref30}{\textbf{[30]}}, \hyperlink{ref32}{\textbf{[32]}}.

Third, the Linear model accentuates the decoupling between prices and profits observed in homogeneous competition. This is most severe in the \textit{DDQN vs.\ DDPG} matchup, where both algorithms suffer substantial losses (DDQN: $\Delta = -0.824$; DDPG: $\Delta = -13.45$). Here, simultaneously learning agents generate a non-stationary environment in which each algorithm's optimal response shifts continuously, preventing stable convergence and inducing destructive experimentation. The resulting instability mirrors known challenges in multi-agent RL with elastic demand and limited state observability \hyperlink{ref11}{\textbf{[11]}}, \hyperlink{ref16}{\textbf{[16]}}, \hyperlink{ref35}{\textbf{[35]}}.

Collectively, the results demonstrate that heterogeneous competition is governed not by inherent tendencies of individual algorithms but by the interaction of their learning mechanisms with market structure. Temporal-difference learners exploit non-adaptive heuristics (e.g., PSO); policy-gradient methods excel against weaker opponents in differentiated markets; but sophisticated learners may mutually destabilize one another in highly elastic settings. Consequently, algorithmic diversity amplifies both strategic asymmetry and the sensitivity of market outcomes to underlying structural features.

\subsection{Demand Shocks, Homogeneous Competition, and Algorithmic Performance}

\subsubsection{Impact on Collusion Indicators}

\begin{figure*}
\centering
\includegraphics[width=0.9\textwidth]{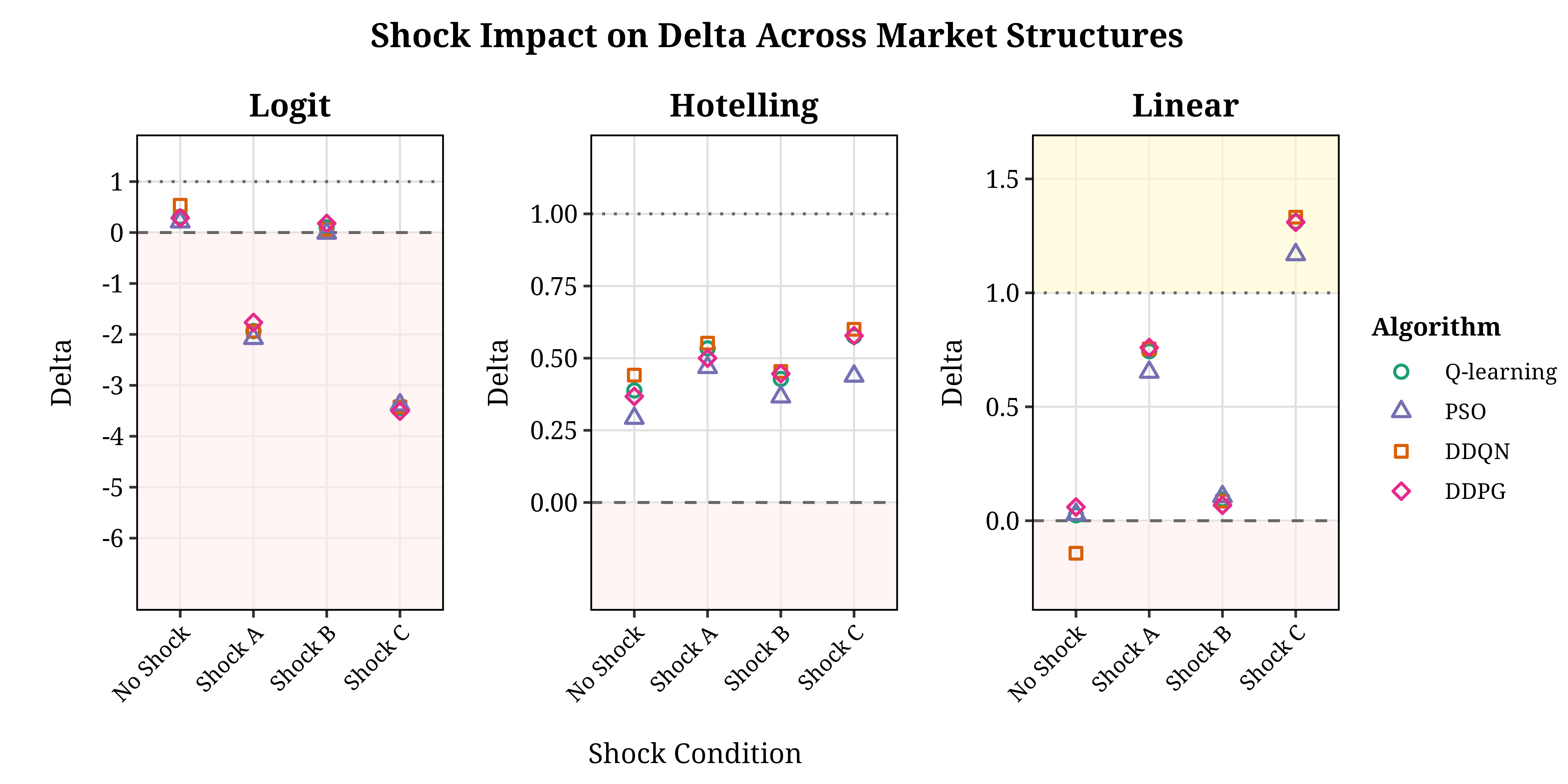}
\caption{Shock Impact on Delta Across Market Structures}
\label{fig:shock_impact}
\end{figure*}

\begin{table*}
\centering
\caption{Impact of Demand Shocks - Logit Model}
\label{tab:Impact of Demand Shocks - Logit Model}
\begin{tabular}{lcccccccc}
\toprule
 & \multicolumn{2}{c}{\textbf{No Shock}} & \multicolumn{2}{c}{\textbf{Scheme A}} & \multicolumn{2}{c}{\textbf{Scheme B}} & \multicolumn{2}{c}{\textbf{Scheme C}} \\
\cmidrule(lr){2-3} \cmidrule(lr){4-5} \cmidrule(lr){6-7} \cmidrule(lr){8-9}
\textbf{Algorithm} & $\mathbf{\Delta}$ & \textbf{RPDI} & $\mathbf{\Delta}$ & \textbf{RPDI} & $\mathbf{\Delta}$ & \textbf{RPDI} & $\mathbf{\Delta}$ & \textbf{RPDI} \\
\midrule
\textbf{Q-Learning} & 0.36 & 0.36 & -1.68 & -0.53 & 0.18 & 0.24 & -2.99 & -0.92 \\
\textbf{DDQN} & 0.42 & 0.41 & -1.49 & -0.45 & 0.23 & 0.29 & -2.81 & -0.86 \\
\textbf{PSO} & 0.33 & 0.22 & -1.90 & -0.78 & 0.12 & 0.08 & -3.46 & -1.23 \\
\textbf{DDPG} & 0.46 & 0.56 & -1.64 & -0.36 & 0.23 & 0.44 & -3.53 & -0.94 \\
\bottomrule
\end{tabular}

\end{table*}

\paragraph{The Logit Model}

The Logit model displays the most severe performance deterioration under demand shocks, a pattern clearly visible in Figure~\ref{fig:shock_impact}, where all algorithms exhibit sharp downward movements in $\Delta$ under Schemes~A and~C. Consistent with these graphical patterns, Table~\ref{tab:Impact of Demand Shocks - Logit Model} shows that Q-Learning records ($\Delta = -2.99$, RPDI = -0.92) under Scheme~C, implying profits 299\% below the shock-adjusted Nash benchmark and prices 92\% below the competitive level. PSO suffers the largest decline ($\Delta = -3.46$, RPDI = -1.23), while DQN and DDPG also incur substantial negative values.

These results arise from the interaction between multiplicative shocks and the exponential form of Logit demand. Because demand is proportional to $\exp((\alpha - p_i + \epsilon_i)/\mu)$, Jensen's inequality \hyperlink{ref39}{\textbf{[39]}} implies that $\mathbb{E}[\exp(\epsilon)] > 1$ whenever shock variance is positive. Expected demand at any fixed price therefore increases under shocks, shifting the profit-maximizing price upward. However, as Figure~\ref{fig:shock_impact} illustrates, algorithms fail to adjust their policies accordingly: they continue pricing near the no-shock Nash equilibrium, generating sub-optimal profits when evaluated against the shock-adjusted benchmark. In contrast, Scheme~B induces only mild distortions -- Q-Learning ($\Delta = 0.18$, RPDI = 0.24) and DDQN ($\Delta = 0.23$, RPDI = 0.29) remain positive -- because its high persistence and low innovation variance enable effective learning. The Logit model thus demonstrates that volatile or high-amplitude shocks can overwhelm algorithmic adaptation, while gradual changes remain tractable.

\begin{table*}
\centering
\caption{Impact of Demand Shocks - Hotelling Model}
\label{tab:Impact of Demand Shocks - Hotelling Model}
\begin{tabular}{lcccccccc}
\toprule
 & \multicolumn{2}{c}{\textbf{No Shock}} & \multicolumn{2}{c}{\textbf{Scheme A}} & \multicolumn{2}{c}{\textbf{Scheme B}} & \multicolumn{2}{c}{\textbf{Scheme C}} \\
\cmidrule(lr){2-3} \cmidrule(lr){4-5} \cmidrule(lr){6-7} \cmidrule(lr){8-9}
\textbf{Algorithm} & $\mathbf{\Delta}$ & \textbf{RPDI} & $\mathbf{\Delta}$ & \textbf{RPDI} & $\mathbf{\Delta}$ & \textbf{RPDI} & $\mathbf{\Delta}$ & \textbf{RPDI} \\
\midrule
\textbf{Q-Learning} & 0.44 & 0.48 & 0.49 & 0.52 & 0.43 & 0.47 & 0.49 & 0.52 \\
\textbf{DDQN} & 0.42 & 0.46 & 0.54 & 0.57 & 0.45 & 0.49 & 0.58 & 0.61 \\
\textbf{PSO} & 0.24 & 0.25 & 0.24 & 0.25 & 0.24 & 0.25 & 0.24 & 0.25 \\
\textbf{DDPG} & 0.51 & 0.52 & 0.50 & 0.54 & 0.45 & 0.48 & 0.65 & 0.70 \\
\bottomrule
\end{tabular}
\end{table*}

\paragraph{The Hotelling Model: Structural Stability Through Spatial Differentiation}

In sharp contrast to the Logit environment, the Hotelling model exhibits striking stability across all shock regimes, a feature clearly represented in Figure~\ref{fig:shock_impact}, where $\Delta$ values remain tightly clustered and consistently positive. Table~\ref{tab:Impact of Demand Shocks - Hotelling Model} confirms this robustness: under Scheme~C, Q-Learning achieves ($\Delta = 0.49$, RPDI = 0.52), nearly identical to its no-shock performance ($\Delta = 0.44$, RPDI = 0.48), while PSO shows complete invariance. This resilience reflects the additive nature of shocks in the Hotelling specification, where disturbances enter independently of prices and therefore do not alter the first-order conditions defining Nash equilibrium. Spatial differentiation further insulates firms by dampening the competitive impact of stochastic fluctuations. As depicted in Figure~\ref{fig:shock_impact_all}, algorithmic performance remains remarkably stable, revealing that market structure -- not learning architecture -- drives shock resistance in differentiated environments.

\begin{table*}
\centering
\caption{Impact of Demand Shocks - Linear Model}
\label{tab:Impact of Demand Shocks - Linear Model}
\begin{tabular}{lcccccccc}
\toprule
 & \multicolumn{2}{c}{\textbf{No Shock}} & \multicolumn{2}{c}{\textbf{Scheme A}} & \multicolumn{2}{c}{\textbf{Scheme B}} & \multicolumn{2}{c}{\textbf{Scheme C}} \\
\cmidrule(lr){2-3} \cmidrule(lr){4-5} \cmidrule(lr){6-7} \cmidrule(lr){8-9}
\textbf{Algorithm} & $\mathbf{\Delta}$ & \textbf{RPDI} & $\mathbf{\Delta}$ & \textbf{RPDI} & $\mathbf{\Delta}$ & \textbf{RPDI} & $\mathbf{\Delta}$ & \textbf{RPDI} \\
\midrule
\textbf{Q-Learning} & 0.04 & 0.49 & 0.74 & 0.54 & 0.10 & 0.49 & 1.31 & 0.53 \\
\textbf{DQN} & 0.05 & 0.54 & 0.77 & 0.63 & 0.11 & 0.57 & 1.35 & 0.64 \\
\textbf{PSO} & 0.01 & 0.05 & 0.60 & 0.05 & 0.07 & 0.05 & 1.11 & 0.05 \\
\textbf{DDPG} & 0.12 & 0.22 & 0.78 & 0.70 & 0.08 & 0.44 & 1.34 & 0.64 \\
\bottomrule
\end{tabular}

\end{table*}

\paragraph{The Linear Model: Elastic Demand and Shock-Induced Profit Inflation}

The Linear model produces an opposite pattern: shocks substantially elevate profit indicators even though prices remain stable. This effect is evident in Figure~\ref{fig:shock_impact}, where all algorithms experience a marked upward shift in $\Delta$ under the more volatile shock schemes. Table~\ref{tab:Impact of Demand Shocks - Linear Model} reports that under Scheme~C, all algorithms exceed $\Delta > 1.0$, with Q-Learning at ($\Delta = 1.31$, RPDI = 0.53) and PSO at ($\Delta = 1.11$, RPDI = 0.05). These gains arise because highly elastic demand imposes severe price penalties in stable environments, but large persistent shocks generate temporary asymmetries in market shares that weaken competitive pressure. Firms can therefore extract higher profits without inducing the substantial quantity losses characteristic of the no-shock equilibrium. As illustrated in Figure~\ref{fig:shock_impact_all}, these improvements are widespread across algorithms, highlighting that elastic demand in combination with persistent shocks can transform uncertainty into systematic profit opportunities.

\paragraph{Cross-Model Synthesis}

\begin{figure*}
\centering
\includegraphics[width=0.9\textwidth]{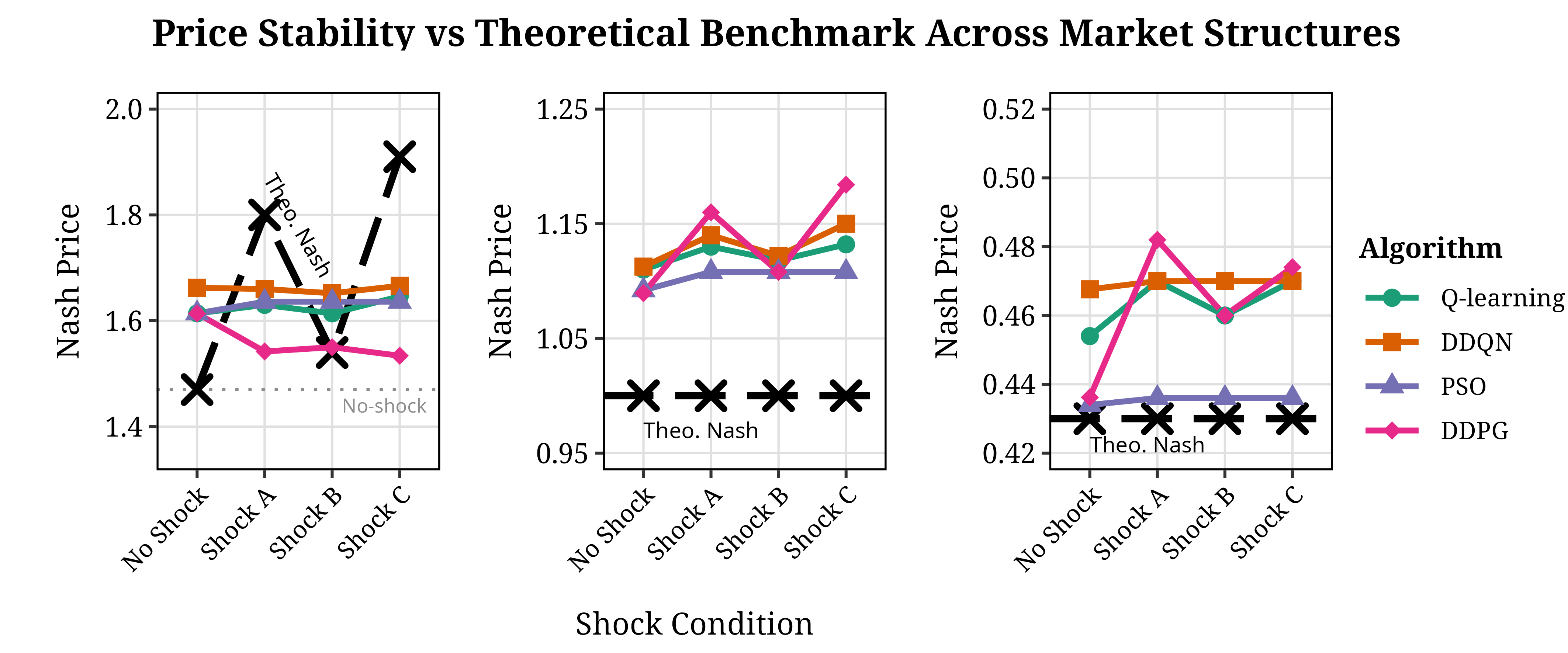}
\caption{Logit Model: Price Stability vs Shifting Benchmark}
\label{fig:price_benchmark}
\end{figure*}

The heterogeneous responses across market structures indicate that algorithmic pricing under uncertainty is shaped fundamentally by the underlying demand specification. In the Logit environment, multiplicative shocks alter the effective choice probabilities through the exponential transformation, generating shifts in the competitive benchmark that learning algorithms fail to track. As illustrated in Figure~\ref{fig:price_benchmark}, even modest increases in shock variance lead to sharp declines in $\Delta$ across all algorithms, reflecting a systematic inability to adapt to the shock-induced displacement of the profit-maximizing price.

By contrast, the Hotelling model exhibits considerable robustness. Spatial differentiation provides a natural buffer against stochastic fluctuations, limiting the extent to which shocks alter relative demand shares. Algorithmic performance therefore remains tightly clustered and largely stable across regimes. In the Linear model, where demand is highly elastic and shocks enter additively, algorithms frequently convert stochastic perturbations into local profit opportunities, yielding performance that is more variable yet consistently non-negative.

Despite these structural contrasts, relative rankings among algorithms show strong within-model consistency. Q-Learning and DQN generally achieve higher profit-based indicators than PSO, while DDPG performs competitively in differentiated settings. This persistence suggests that learning architecture exerts a stable influence on relative performance even when absolute performance shifts markedly. The pattern is particularly clear in the Logit model, where average prices remain remarkably stable despite large movements in $\Delta$ and RPDI values; Table~\ref{tab:Average Prices Remain Stable Despite Shock-Induced Performance Changes - Logit Model} shows that prices change little across shock regimes even as the Nash benchmark shifts substantially.

Taken together, these results imply that market structure, rather than algorithm class, is the primary determinant of shock sensitivity. This has direct implications for competitive behavior and regulatory oversight: identical algorithms may exhibit drastically different vulnerability profiles depending on the demand environment in which they operate.

\begin{table*}
\centering
\caption{Average Prices Remain Stable Despite Shock-Induced Performance Changes - Logit Model}
\label{tab:Average Prices Remain Stable Despite Shock-Induced Performance Changes - Logit Model}
\begin{tabular}{lcccc}
\toprule
\textbf{Algorithm} & \textbf{No Shock} & \textbf{Scheme A} & \textbf{Scheme B} & \textbf{Scheme C} \\
\midrule
\textbf{Q-Learning} & 1.63 & 1.65 & 1.64 & 1.66 \\
\textbf{DDQN} & 1.66 & 1.67 & 1.66 & 1.67 \\
\textbf{PSO} & 1.57 & 1.57 & 1.57 & 1.57 \\
\textbf{DDPG} & 1.72 & 1.70 & 1.72 & 1.65 \\
\textbf{Nash (Shock-Adj.)} & 1.47 & 1.80 & 1.54 & 1.91 \\
\bottomrule
\end{tabular}

\end{table*}

\subsubsection{Shock Scheme Comparison}

Table~\ref{tab:Shock Scheme Effects - Performance Change from Baseline} summarizes the change in the profit-based indicator relative to the no-shock baseline, revealing clear differences in sensitivity across market structures. The Logit model exhibits the largest performance losses, with all algorithms experiencing severe declines under Schemes~A and~C (e.g., Q-Learning: $-2.32$ and $-3.34$), confirming the model's vulnerability to shock-induced benchmark shifts. The Hotelling model shows near-zero deviations across all schemes, indicating strong structural stability and minimal disruption to learned strategies. In contrast, the Linear model consistently produces positive performance gains, particularly under Scheme~C where improvements exceed $+1.0$ for all algorithms. These patterns demonstrate that the direction and magnitude of shock effects depend primarily on market structure rather than algorithmic class.

\begin{table*}
\centering
\caption{Shock Scheme Effects - Performance Change from Baseline ($\Delta\Delta$)}
\label{tab:Shock Scheme Effects - Performance Change from Baseline}
\begin{tabular}{lccccccccc}
\toprule
 & \textbf{Log A} & \textbf{Log B} & \textbf{Log C} & \textbf{Hot A} & \textbf{Hot B} & \textbf{Hot C} & \textbf{Lin A} & \textbf{Lin B} & \textbf{Lin C} \\
\midrule
\textbf{Q-Learning} & -2.32 & -0.16 & -3.34 & +0.08 & +0.01 & +0.07 & +0.70 & +0.07 & +1.28 \\
\textbf{DDQN} & -1.91 & -0.18 & -3.21 & +0.12 & +0.02 & +0.15 & +0.72 & +0.06 & +1.31 \\
\textbf{PSO} & -2.21 & -0.20 & -3.75 & +0.01 & +0.00 & +0.01 & +0.59 & +0.06 & +1.09 \\
\textbf{DDPG} & -2.04 & -0.21 & -3.83 & +0.06 & -0.03 & +0.10 & +0.66 & -0.01 & +1.20 \\
\bottomrule
\end{tabular}

\end{table*}

\subsection{Heterogeneous Competition Under Demand Shocks}

\begin{table*}
\centering
\caption{Key Matchups Under Scheme C - All Markets}
\label{tab:Key Matchups Under Scheme C - All Markets}
\begin{tabular}{lcccccc}
\toprule
 & \multicolumn{2}{c}{\textbf{Logit Model}} & \multicolumn{2}{c}{\textbf{Hotelling Model}} & \multicolumn{2}{c}{\textbf{Linear Model}} \\
\cmidrule(lr){2-3} \cmidrule(lr){4-5} \cmidrule(lr){6-7}
\textbf{Matchup} & $\mathbf{\Delta}$ & \textbf{RPDI} & $\mathbf{\Delta}$ & \textbf{RPDI} & $\mathbf{\Delta}$ & \textbf{RPDI} \\
\midrule
\textbf{Q-Learning vs PSO} & -2.26 & -0.99 & 0.56 & 0.53 & 1.27 & 0.52 \\
 & -2.85 & -0.81 & 0.53 & 0.57 & 1.19 & 0.14 \\
\midrule
\textbf{Q-Learning vs DDQN} & -2.87 & -0.81 & 0.57 & 0.54 & 1.33 & 0.53 \\
 & -2.63 & -0.86 & 0.51 & 0.59 & 1.32 & 0.63 \\
\midrule
\textbf{Q-Learning vs DDPG} & -6.01 & -1.13 & 0.78 & 0.53 & 1.34 & 0.52 \\
 & -3.78 & -1.72 & 0.53 & 0.86 & 1.33 & 0.70 \\
\midrule
\textbf{PSO vs DDQN} & -2.30 & -0.81 & 0.60 & 0.58 & 1.22 & 0.14 \\
 & -2.50 & -0.88 & 0.54 & 0.59 & 1.28 & 0.64 \\
\midrule
\textbf{PSO vs DDPG} & -4.85 & -0.82 & 0.58 & 0.55 & 1.21 & 0.13 \\
 & -3.42 & -1.43 & 0.47 & 0.57 & 1.23 & 0.54 \\
\midrule
\textbf{DQN vs DDPG} & -6.36 & -0.94 & 0.80 & 0.60 & 1.36 & 0.63 \\
 & -3.28 & -1.73 & 0.59 & 0.86 & 1.32 & 0.63 \\
\bottomrule
\end{tabular}
\end{table*}

\subsubsection{Cross-Market Patterns and Algorithm-Specific Responses}

Cross-market comparisons under Scheme~C reveal pronounced structural differences in how heterogeneous algorithmic pairings respond to persistent demand volatility. In the Logit model, severe performance deterioration persists across all matchups in Table~\ref{tab:Key Matchups Under Scheme C - All Markets}. As the price-benchmark gap widens under shocks (illustrated in Figure~\ref{fig:price_benchmark}), every algorithm records strongly negative values of both $\Delta$ and RPDI. Within this generally adverse environment, modest asymmetries emerge: in the Q-Learning--PSO pairing, Q-Learning exhibits comparatively smaller losses ($\Delta = -2.26$, RPDI $= -0.99$) than PSO ($\Delta = -2.85$, RPDI $= -0.81$), reflecting the stabilizing effect of temporal-difference learning relative to PSO's memoryless search \hyperlink{ref37}{\textbf{[37]}}, \hyperlink{ref38}{\textbf{[38]}}. A more substantial divergence is observed in the DDQN--DDPG matchup, where DDQN reaches ($\Delta = -6.36$, RPDI $= -0.94$) while DDPG attains ($\Delta = -3.28$, RPDI $= -1.73$), underscoring the vulnerability of discrete-action deep Q-learning to non-stationary, shock-driven environments \hyperlink{ref29}{\textbf{[29]}}, \hyperlink{ref35}{\textbf{[35]}}.

By contrast, the Hotelling model maintains robust, shock-resistant behavior across heterogeneous pairings. Values remain tightly clustered and positive in Table~\ref{tab:Key Matchups Under Scheme C - All Markets}, consistent with the minimal deviation from theoretical benchmarks shown in Figure~\ref{fig:price_benchmark}. In Q-Learning--DDQN, both algorithms sustain collusive alignment ($\Delta = 0.57$, RPDI = 0.54; $\Delta = 0.51$, RPDI = 0.59), suggesting that spatial differentiation dampens volatility and suppresses opportunities for strategic exploitation. The additive nature of shocks in this structure avoids the benchmark shifts that destabilize learning in Logit markets, allowing each algorithm to preserve coordinated behavior.

The Linear model continues to display the shock-induced profit inflation observed in homogeneous competition, with all heterogeneous pairings in Table~\ref{tab:Key Matchups Under Scheme C - All Markets} recording $\Delta > 1.0$. Notably, Figure~\ref{fig:price_benchmark} shows relatively stable pricing even as profits rise, highlighting the decoupling between price and profit under elastic demand. In Q-Learning--PSO, both algorithms exceed $\Delta = 1.2$, though their mechanisms differ: Q-Learning raises prices (RPDI = 0.52), whereas PSO maintains near-competitive pricing (RPDI = 0.14) while benefiting disproportionately from favorable shock realizations. The elastic structure amplifies temporary demand asymmetries, enabling both sophisticated and simple algorithms to capture supra-competitive profits regardless of strategic capacity.

Collectively, these results demonstrate that market structure remains the dominant determinant of performance under volatility, shaping not only the magnitude of shock impacts but also the direction of competitive asymmetries across heterogeneous algorithmic pairings.

\subsubsection{Consumer Welfare Implications}

\begin{table*}
\centering
\caption{Consumer Surplus Change Relative to Nash Equilibrium (\%) - Logit Model}
\label{tab:Consumer Surplus Change Relative to Nash Equilibrium - Logit Model}
\begin{tabular}{lcccc}
\toprule
\textbf{Matchup} & \textbf{No Shock} & \textbf{Scheme A} & \textbf{Scheme B} & \textbf{Scheme C} \\
\midrule
\textbf{Q-Learning vs Q-Learning} & -20.6\% & +31.3\% & -13.3\% & +62.1\% \\
\textbf{DDQN vs DDQN} & -24.3\% & +26.1\% & -16.7\% & +58.1\% \\
\textbf{PSO vs PSO} & -13.0\% & +47.4\% & -4.2\% & +84.8\% \\
\textbf{DDPG vs DDPG} & -31.6\% & +20.9\% & -24.1\% & +64.8\% \\
\bottomrule
\end{tabular}

\end{table*}

Consumer welfare responses to demand shocks in the Logit model reveal a striking reversal of outcomes across regimes. Under no-shock conditions, all algorithms reduce consumer surplus relative to Nash equilibrium, with losses ranging from $-13.0\%$ for PSO to $-31.6\%$ for DDPG, reflecting their tendency to maintain supra-competitive pricing. Under Scheme~C, this pattern flips: Q-Learning generates a $+62.1\%$ surplus gain, PSO $+84.8\%$, and DDPG $+64.8\%$, as shown in Table~\ref{tab:Consumer Surplus Change Relative to Nash Equilibrium - Logit Model}. These gains arise not from more competitive behavior but from algorithms failing to adjust prices upward when the shock-adjusted optimal price increases, thereby unintentionally benefiting consumers. Scheme~B produces moderate welfare losses consistent with partial adaptation to gradual drift, while Scheme~A yields positive but less pronounced gains due to its short-lived volatility. Overall, the results indicate that demand shocks enhance consumer surplus primarily by disrupting algorithmic coordination, suggesting that such welfare improvements are incidental and may not persist as algorithms become more adaptive.

\subsection{Synthesis and Key Findings}

Across all environments, the interaction between algorithmic design and market structure emerges as the primary determinant of pricing outcomes. Under stable demand, reinforcement learning methods such as Q-Learning, DQN, and DDPG consistently generate elevated collusion indicators in the Logit and Hotelling models, with DDPG achieving the highest levels of supra-competitive behavior. PSO, by contrast, exhibits near-competitive outcomes across all markets, reflecting its heuristic and memoryless optimization structure. The Linear model displays a distinct dynamic: algorithms sustain elevated prices without corresponding profit gains, producing a systematic decoupling between price elevation and profitability that reduces consumer surplus without providing substantial benefits to firms.

Demand shocks amplify these structural differences. The Logit model proves highly vulnerable, with Scheme~C generating severe performance collapses for all algorithms, driven by multiplicative shock effects that shift the profit-maximizing benchmark beyond the algorithms' adaptive capacity. The Hotelling model remains robust, with spatial differentiation buffering the impact of shocks and maintaining performance near baseline across both homogeneous and heterogeneous matchups. In the Linear model, shocks invert the usual competitive dynamics: persistent disturbances generate substantial profit inflation, as transient demand asymmetries weaken competitive pressure while prices remain stable. Despite these absolute shifts, relative algorithmic rankings remain consistent within each market, indicating that underlying learning architectures retain their comparative strengths even under significant environmental volatility.

\section{Conclusion}

This study demonstrates that algorithmic pricing outcomes depend fundamentally on the interaction between learning architecture, market structure, and demand uncertainty. Reinforcement learning algorithms sustain supra-competitive prices in stable Logit and Hotelling environments, while PSO remains near-competitive. Demand shocks amplify structural differences: the Logit model exhibits severe instability under persistent shocks, Hotelling remains largely unaffected, and the Linear model converts volatility into profit gains despite stable pricing. These patterns persist in both homogeneous and heterogeneous competition, indicating that market structure -- not algorithmic diversity -- primarily governs coordination tendencies and shock vulnerability. Consumer welfare effects reflect this asymmetry, with shocks disrupting coordination in Logit but leaving differentiated markets largely unchanged.

Several limitations warrant caution. The analysis focuses on duopolies, fixed hyperparameters, and stationary learning rules, which may not capture the full range of real-world strategic adaptation. Algorithms with explicit opponent modeling, dynamic state augmentation, or meta-learning could exhibit materially different behavior. Moreover, the study abstracts from inventory constraints, multi-product interactions, and platform ranking mechanisms, all of which may influence algorithmic pricing dynamics. Future work should extend these insights to richer competitive environments and explore regulatory implications in markets where algorithmic learning interacts with complex demand structures.

\section*{References}

\begin{enumerate}[label={[\arabic*]}]

\item \hypertarget{ref1}{} J.~M.~Sanchez-Cartas and E.~Katsamakas, 
"Artificial Intelligence, Algorithmic Competition and Market Structures," 
\emph{IEEE Access}, vol.~10, pp.~10575--10584, 2022.\\
DOI: \url{10.1109/ACCESS.2022.3144390}

\item \hypertarget{ref2}{} E.~Calvano, G.~Calzolari, V.~Denicolò, and S.~Pastorello, 
"Artificial Intelligence, Algorithmic Pricing, and Collusion," 
\emph{American Economic Review}, vol.~110, no.~10, pp.~3267--3297, Oct.~2020.\\
DOI: \url{10.1257/aer.20190623}

\item \hypertarget{ref3}{} J.~M.~Sánchez-Cartas, A.~Tejero, and G.~León, 
"Algorithmic Pricing and Price Gouging. Consequences of High-Impact, Low Probability Events," 
\emph{Sustainability}, vol.~13, no.~5, p.~2542, Feb.~2021.\\
DOI: \url{10.3390/su13052542}

\item \hypertarget{ref4}{} U.~Schwalbe, 
"ALGORITHMS, MACHINE LEARNING, AND COLLUSION," 
\emph{Journal of Competition Law \& Economics}, vol.~14, no.~4, pp.~568--607, Dec.~2018.\\
DOI: \url{10.1093/joclec/nhz004}

\item \hypertarget{ref5}{} T.~Klein, 
"Autonomous algorithmic collusion: Q‐learning under sequential pricing," 
\emph{Rand Journal of Economics}, vol.~52, no.~3, pp.~538--558, Sep.~2021.\\
DOI: \url{10.1111/1756-2171.12383}

\item \hypertarget{ref6}{} E.~Calvano, G.~Calzolari, V.~Denicolò, and S.~Pastorello, 
"Algorithmic collusion with imperfect monitoring," 
\emph{International Journal of Industrial Organization}, vol.~79, p.~102712, Dec.~2021.\\
DOI: \url{10.1016/j.ijindorg.2021.102712}

\item \hypertarget{ref7}{} C.~J.~C.~H.~Watkins and P.~Dayan, 
"Q-Learning," 1992.

\item \hypertarget{ref8}{} F.~Lange, L.~Dreessen, and R.~Schlosser, 
"Reinforcement learning versus data-driven dynamic programming: a comparison for finite horizon dynamic pricing markets," 
\emph{Journal of Revenue and Pricing Management}, Dec.~2025.\\
DOI: \url{10.1057/s41272-025-00519-8}

\item \hypertarget{ref9}{} P.~Belleflamme and M.~Peitz, 
\emph{Industrial Organization: Markets and Strategies}. 
Cambridge University Press, 2018.

\item \hypertarget{ref10}{} J.~E.~Harrington, 
"The Effect of Outsourcing Pricing Algorithms on Market Competition," 2021.\\
Available: \url{https://ssrn.com/abstract=3798847}

\item \hypertarget{ref11}{} A.~Kastius and R.~Schlosser, 
"Dynamic pricing under competition using reinforcement learning," 
\emph{Journal of Revenue and Pricing Management}, vol.~21, no.~1, pp.~50--63, Feb.~2022.\\
DOI: \url{10.1057/s41272-021-00285-3}

\item \hypertarget{ref12}{} V.~Mnih et al., 
"Human-level control through deep reinforcement learning," 
\emph{Nature}, vol.~518, no.~7540, pp.~529--533, Feb.~2015.\\
DOI: \url{10.1038/nature14236}

\item \hypertarget{ref13}{} M.~Hessel et al., 
"Rainbow: Combining Improvements in Deep Reinforcement Learning."\\
Available: \url{https://www.aaai.org}

\item \hypertarget{ref14}{} T.~P.~Lillicrap et al., 
"Continuous control with deep reinforcement learning," Jul.~2019.\\
Available: \url{http://arxiv.org/abs/1509.02971}

\item \hypertarget{ref15}{} D.~Liu, W.~Wang, L.~Wang, H.~Jia, and M.~Shi, 
"Dynamic Pricing Strategy of Electric Vehicle Aggregators Based on DDPG Reinforcement Learning Algorithm," 
\emph{IEEE Access}, vol.~9, pp.~21556--21566, 2021.\\
DOI: \url{10.1109/ACCESS.2021.3055517}

\item \hypertarget{ref16}{} E.~H.~Sumiea et al., 
"Deep deterministic policy gradient algorithm: A systematic review," 
\emph{Heliyon}, vol.~10, no.~9, May 2024.\\
DOI: \url{10.1016/j.heliyon.2024.e30697}

\item \hypertarget{ref17}{} A.~V.~Den~Boer, 
"Dynamic Pricing and Learning: Historical Origins, Current Research, and New Directions," 2015.\\
Available: \url{https://ssrn.com/abstract=2334429}

\item \hypertarget{ref18}{} W.~Elmaghraby and P.~Keskinocak, 
"Dynamic Pricing in the Presence of Inventory Considerations: Research Overview, Current Practices, and Future Directions," 2003.

\item \hypertarget{ref19}{} D.~D.~Cox and I.~Llatas, 
"Maximum Likelihood Type Estimation for Nearly Nonstationary Autoregressive Time Series," 1991.\\
Available: \url{http://www.jstor.org/stable/2241941}

\item \hypertarget{ref20}{} Z.~Lu and K.~Shimizu, 
"Estimating Discrete Choice Demand Models with Sparse Market-Product Shocks," Jul.~2025.\\
Available: \url{http://arxiv.org/abs/2501.02381}

\item \hypertarget{ref21}{} D.~W.~K.~Andrews, 
"Exactly Median-Unbiased Estimation of First Order Autoregressive/Unit Root Models," 1993.

\item \hypertarget{ref22}{} S.~Deng, M.~Schiffer, and M.~Bichler, 
"Algorithmic Collusion in Dynamic Pricing with Deep Reinforcement Learning," Jun.~2024.\\
Available: \url{http://arxiv.org/abs/2406.02437}

\item \hypertarget{ref23}{} K.~E.~Train, 
\emph{Discrete Choice Methods with Simulation}, 2002.

\item \hypertarget{ref24}{} J.~P.~Johnson et al., 
"Platform Design When Sellers Use Pricing Algorithms," 2022.

\item \hypertarget{ref25}{} E.~J.~Anderson, 
"On the existence of supply function equilibria," 
\emph{Mathematical Programming}, vol.~140, no.~2, pp.~323--349, Sep.~2013.\\
DOI: \url{10.1007/s10107-013-0691-7}

\item \hypertarget{ref26}{} R.~Anupindi and L.~Jiang, 
"Capacity investment under postponement strategies, market competition, and demand uncertainty," 
\emph{Management Science}, vol.~54, no.~11, pp.~1876--1890, Nov.~2008.\\
DOI: \url{10.1287/mnsc.1080.0940}

\item \hypertarget{ref27}{} R.~Balvers and L.~Szerb, 
"Location in the Hotelling duopoly model with demand uncertainty," 1996.

\item \hypertarget{ref28}{} R.~Shi, J.~Zhang, and J.~Ru, 
"Impacts of power structure on supply chains with uncertain demand," 
\emph{Production and Operations Management}, vol.~22, no.~5, pp.~1232--1249, Sep.~2013.\\
DOI: \url{10.1111/poms.12002}

\item \hypertarget{ref29}{} Y.~Ye, D.~Qiu, M.~Sun, D.~Papadaskalopoulos, and G.~Strbac, 
"Deep Reinforcement Learning for Strategic Bidding in Electricity Markets," 
\emph{IEEE Transactions on Smart Grid}, vol.~11, no.~2, pp.~1343--1355, Mar.~2020.\\
DOI: \url{10.1109/TSG.2019.2936142}

\item \hypertarget{ref30}{} Q.~Zhao, G.~Mao, and S.~Wang, 
"Balancing Profit and Cultural Heritage: Multi-Objective Dynamic Pricing for Hanfu Using Deep Deterministic Policy Gradient," 
\emph{IEEE Access}, vol.~13, pp.~1--15, 2025.\\
DOI: \url{10.1109/ACCESS.2025.3526685}

\item \hypertarget{ref31}{} A.~G.~Gad, 
"Particle Swarm Optimization Algorithm and Its Applications: A Systematic Review," 
\emph{Archives of Computational Methods in Engineering}, vol.~29, no.~5, pp.~2531--2561, Aug.~2022.\\
DOI: \url{10.1007/s11831-021-09694-4}

\item \hypertarget{ref32}{} D.~Qiu, Y.~Ye, D.~Papadaskalopoulos, and G.~Strbac, 
"A Deep Reinforcement Learning Method for Pricing Electric Vehicles With Discrete Charging Levels," 
\emph{IEEE Transactions on Industry Applications}, vol.~56, no.~5, pp.~5901--5912, Sep.~2020.\\
DOI: \url{10.1109/TIA.2020.2984618}

\item \hypertarget{ref33}{} J.~Pérez and H.~López-Ospina, 
"Competitive Pricing for Multiple Market Segments Considering Consumers' Willingness to Pay," 
\emph{Mathematics}, vol.~10, no.~19, p.~3600, Oct.~2022.\\
DOI: \url{10.3390/math10193600}

\item \hypertarget{ref34}{} S.~Deng, M.~Schiffer, and M.~Bichler, 
"Exploring Competitive and Collusive Behaviors in Algorithmic Pricing with Deep Reinforcement Learning," 
Mar.~2025.\\
Available: \url{http://arxiv.org/abs/2503.11270}

\item \hypertarget{ref35}{} J.~Sun, Z.~Wang, Z.~Qiao, and X.~Li, 
"Dynamic Pricing Model for E-commerce Products Based on DDQN," 
\emph{Journal of Comprehensive Business Administration Research}, vol.~1, no.~3, pp.~171--178, May~2024.\\
DOI: \url{10.47852/bonviewjcbar42022770}

\item \hypertarget{ref36}{} H.~Weng, Y.~Hu, M.~Liang, J.~Xi, and B.~Yin, 
"Optimizing bidding strategy in electricity market based on graph convolutional neural network and deep reinforcement learning," 
\emph{Applied Energy}, vol.~380, p.~124978, Feb.~2025.\\
DOI: \url{10.1016/j.apenergy.2024.124978}

\item \hypertarget{ref37}{} J.~Zheng, Y.~Gan, Y.~Liang, Q.~Jiang, and J.~Chang, 
"Joint Strategy of Dynamic Ordering and Pricing for Competing Perishables with Q-Learning Algorithm," 
\emph{Wireless Communications and Mobile Computing}, vol.~2021, 2021.\\
DOI: \url{10.1155/2021/6643195}

\item \hypertarget{ref38}{} P.~Famil Alamdar and A.~Seifi, 
"A deep Q-learning approach to optimize ordering and dynamic pricing decisions in the presence of strategic customers," 
\emph{International Journal of Production Economics}, vol.~269, p.~109154, Mar.~2024.\\
DOI: \url{10.1016/j.ijpe.2024.109154}

\item \hypertarget{ref39}{} K. E. Train, ``Discrete Choice Methods with Simulation,'' 2002.
\end{enumerate}

\end{document}